\documentclass[letterpaper, 10 pt, conference]{ieeeconf}
\IEEEoverridecommandlockouts
\usepackage{graphicx}
\usepackage{amsmath}
\usepackage{cite}
\usepackage{url}
\usepackage{stfloats}
\usepackage{amssymb}
\usepackage{booktabs}
\usepackage{hyperref}

\title{\LARGE \bf
LeVR: A Modular VR Teleoperation Framework for Imitation Learning in Dexterous Manipulation
}

\author{Zhengyang Kris Weng$^{\ast,\ddagger}$,
        Matthew Elwin$^{\ast}$,
        Han Liu$^{\dagger}$%
\thanks{$^{\ast}$Center for Robotics and Biosystems, Northwestern University, Evanston, IL, USA.}%
\thanks{$^{\dagger}$MAGICS Lab, Northwestern University, Evanston, IL, USA.}%
\thanks{$^{\ddagger}$Email: \href{mailto:zkweng@u.northwestern.edu}{zkweng@u.northwestern.edu}}%
}

\begin{document}

\maketitle
\thispagestyle{empty}
\pagestyle{empty}


\begin{abstract}
We introduce LeVR, a modular software framework designed to bridge two critical gaps in robotic imitation learning. First, it provides robust and intuitive virtual reality (VR) teleoperation for data collection using robot arms paired with dexterous hands, addressing a common limitation in existing systems. Second, it natively integrates with the powerful LeRobot imitation learning (IL) framework, enabling the use of VR-based teleoperation data and streamlining the demonstration collection process. To demonstrate LeVR, we release LeFranX, an open-source implementation for the Franka FER arm and RobotEra XHand, two widely used research platforms. LeFranX delivers a seamless, end-to-end workflow from data collection to real-world policy deployment. We validate our system by collecting a public dataset of 100 expert demonstrations and use it to successfully fine-tune state-of-the-art visuomotor policies. We provide our open-source framework, implementation\footnote{Code repo: \url{https://github.com/wengmister/LeFranX}}, and dataset\footnote{Dataset:\,\url{https://huggingface.co/collections/wengmister}}
 to accelerate IL research for the robotics community.
\end{abstract}


\section{Introduction}

Imitation learning (IL) has shown strong promise for teaching robots dexterous manipulation, but its success depends on access to large-scale expert data \cite{underactuated}. Virtual reality (VR) teleoperation has emerged as a compelling tool for collecting such data, yet its broader use is limited by two key challenges. First, most existing VR systems provide only limited support for multi-fingered dexterous hands, constraining the range of skills that can be demonstrated. Second, while powerful IL frameworks such as LeRobot\cite{cadene2024lerobot} offer convenient pipelines for training, they do not natively support VR-based demonstration data, creating a gap that makes it difficult to directly leverage teleoperation for policy learning. Bridging this gap typically requires additional integration code and custom interfaces, which can add overhead and slow adoption in practice.

\begin{figure}[t!]
    \centering
    \includegraphics[width=0.45\textwidth]{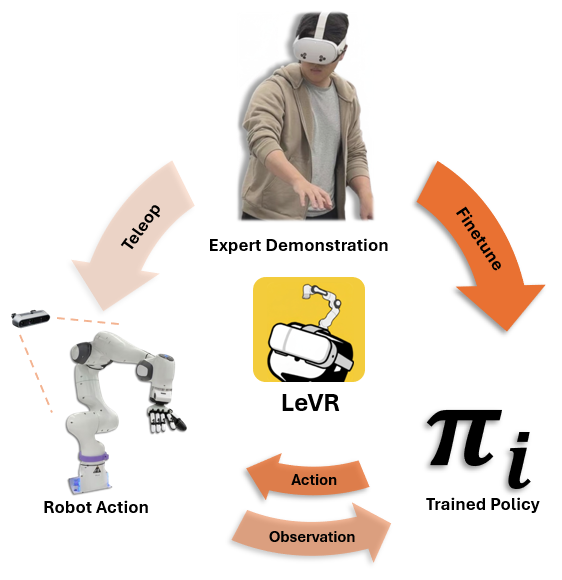}
    \caption{Overview of the LeVR framework, which integrates VR dexterous teleoperation, imitation learning policy fine-tuning ($\pi_i$), and robot action execution into a unified workflow built upon the LeRobot system\cite{cadene2024lerobot}.}
    \label{fig:system_overview}
\end{figure}

To handle the above two challenges, we introduce LeVR, a modular software framework designed to solve both problems. LeVR provides intuitive VR control for dexterous hands and automates the data pipeline for IL frameworks. We develop and release LeFranX, an open-source LeVR implementation for the Franka FER arm\cite{franka_emika_robot} and RobotEra XHand\cite{robotera_xhand1_2024} that provides a complete workflow from data collection to policy deployment. 
Built as an extension for LeRobot, our system combines a low-cost VR interface with robust retargeting algorithms \cite{handa2020dexpilot,qin2023anyteleop} to enable intuitive control. An overview of the LeVR framework is shown in Figure \ref{fig:system_overview}. By providing a principled architecture and a ready-to-use implementation, our work aims to lower the barrier to entry for researchers to explore, benchmark, and share robot learning workflows.

The contributions of this work are: (i) A modular and generalizable VR teleoperation framework that is designed for integration with robot learning pipelines. (ii) An open-source implementation, LeFranX, that instantiates the framework on a Franka FER robot with a RobotEra XHand, including retargeting and high-frequency control. (iii) A public dataset of teleoperated manipulation demonstrations for common manipulation tasks, formatted for use with LeRobot. (iv) Experimental validation that demonstrates the framework's low-latency teleoperation and the efficacy of the collected demonstrations in improving policy performance.

\begin{figure*}[t] 
    \centering
    \includegraphics[width=\textwidth]{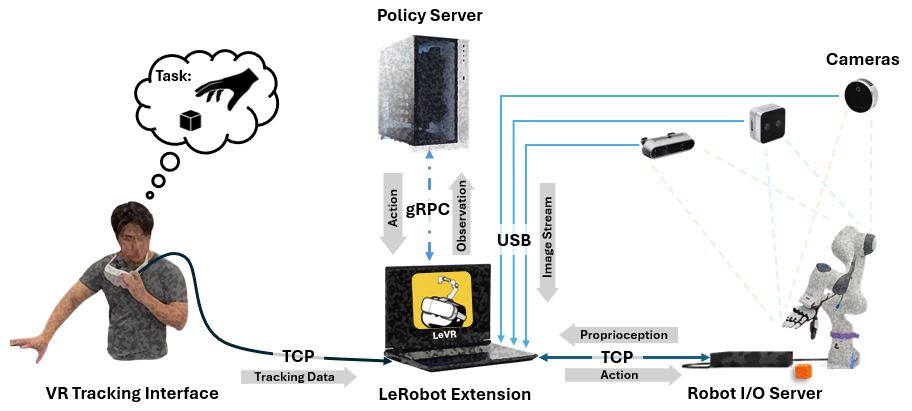}
    \caption{Overview of the system architecture: During demonstration collection, VR interface streams tracking data to the LeRobot extension, and cameras provide image streams. The system receives proprioceptive observations from the robot while sending actions to it. During policy rollout, the system runs inference on the trained checkpoint instead of receiving operator input.}
    \label{fig:system_architecture}
\end{figure*}

\section{Related Work}

Imitation learning (IL) and its variant, behavior cloning (BC), are powerful paradigms for teaching robots complex skills \cite{underactuated}. By learning from human demonstrations, IL avoids the need for manually engineered primitives tied to specific robot kinematics \cite{ARGALL2009469}. A successful IL pipeline comprises two core components: (1) an effective method for collecting expert demonstration data, and (2) a robust framework for training and deploying policies derived from this data. In the policy training stage, researchers often use powerful visuomotor architectures like Action Chunking with Transformers (ACT) \cite{zhao2023learning} or Diffusion Policy (DP) \cite{chi2023diffusionpolicy} to map robot observations to actions.

The first component, data collection, is critical as IL performance is sensitive to data quality and quantity\cite{ARGALL2009469}. Researchers have explored a wide spectrum of teleoperation methods. High-fidelity approaches include kinesthetic teaching \cite{billard2008robot} and full-arm or hand-specific exoskeletons \cite{fang2024airexo, toedtheide2023force, zhang2025doglove, li2020mobile}. While providing natural control, these systems are often expensive, cumbersome, and require the operator to be co-located with the robot. Other systems lower the cost barrier with dedicated hardware like leader-follower arms \cite{zhao2023learning}, tangible replicas \cite{wu2023gello}, or joysticks \cite{liu2023libero}. Alternatively, some methods eliminate custom hardware entirely, using commodity sensors like smartphone IMUs \cite{mandlekar2019roboturk} or RGB cameras \cite{sivakumar2022robotic, qin2023anyteleop, handa2020dexpilot}, but can suffer from noise and imprecision. Among these solutions, VR teleoperation has emerged as a cost-effective sweet spot. Consumer-grade systems like the Meta Quest \cite{MetaQuest3}, Apple Vision Pro \cite{AppleVisionProSpecs}, and HTC Vive Pro \cite{HTCVivePro2Specs} provide high-fidelity, 6-DOF tracking suitable for rich data collection \cite{gharaybeh2019telerobotic, george2023openvr}. Despite recent advances in VR teleoperation research \cite{buckley2024dexterous, iyer2024open, ding2024bunnyvisionprorealtimebimanualdexterous}, a critical challenge remains: most systems are not "friendly" to multi-fingered dexterous hands. The difficulty of retargeting human motion to robot kinematics \cite{arunachalam2022dexterous} leads many implementations to support only simple gripper commands \cite{zhang2018deepimitationlearningcomplex, george2023openvr} or direct joint mappings that perform poorly for kinematically dissimilar hands \cite{iyer2024open}.

The second component of the IL pipeline is policy training. LeRobot\cite{cadene2024lerobot} has recently become a popular and standardized framework for this purpose, offering robust implementations of state-of-the-art policies. However, its officially supported teleoperation solutions are primarily designed for physical hardware, such as leader–follower setups or exoskeletons. At present, the framework does not natively accommodate VR-based teleoperation data, which has become increasingly common. As a result, researchers often need to implement additional software layers to adapt VR-collected demonstrations into the formats required by LeRobot—an integration step that can add substantial engineering overhead.

Our work, LeVR, directly addresses these two challenges. We provide a unified framework that (1) enables intuitive and robust teleoperation for dexterous hands using commodity VR hardware, and (2) offers seamless, out-of-the-box integration with the LeRobot policy training ecosystem.


\section{LeVR System Architecture}

The framework consists of a modular architecture that separates general teleoperation logic from robot-specific extensions. This principled design facilitates easy adaptation to new hardware, requiring only small, well-defined modifications to the extensions for any given embodiment (Fig.~\ref{fig:system_architecture}).


\subsection{Hand-tracking VR Interface}

\begin{figure}[b]
    \centering
    \includegraphics[width=0.475\textwidth]{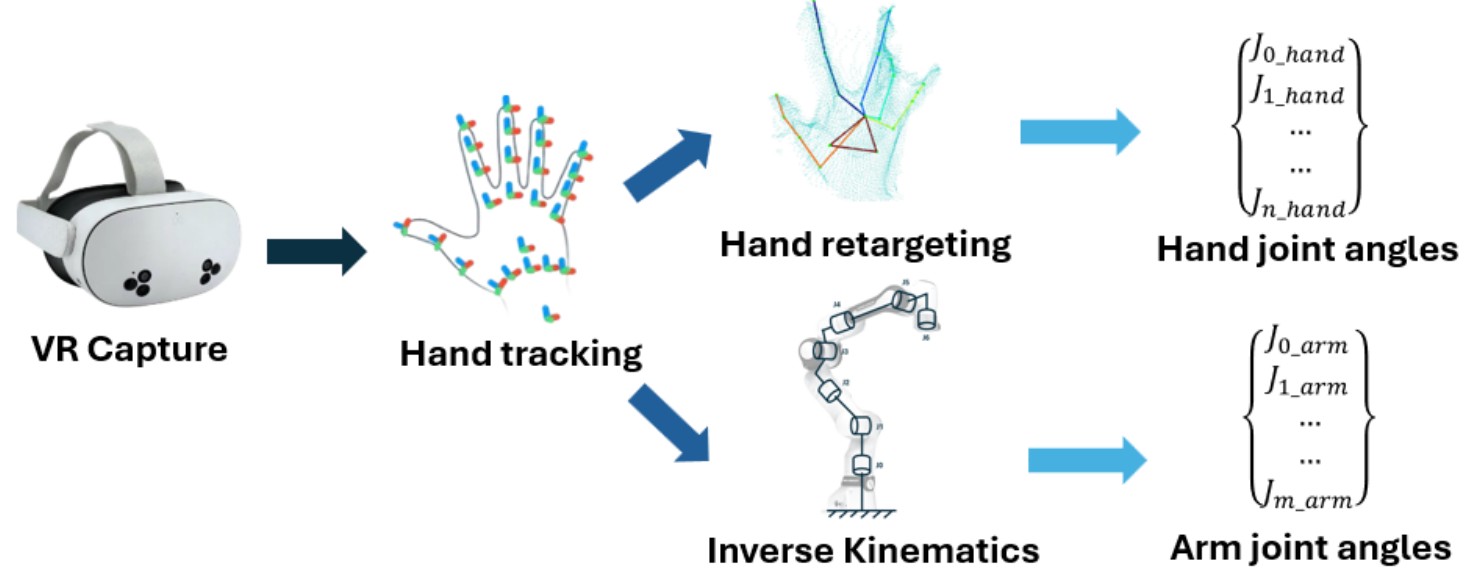}
    \caption{Teleoperation pipeline: VR hand tracking signal is processed and retargeted in the arm branch and hand branch simultaneously.}
    \label{fig:teleop_pipeline}
\end{figure}

The VR interface captures hand motion using the Meta Quest's built-in kinematic tracking via the OpenXR Hand API\cite{KhronosOpenXR}, and is applicable to all single-arm single-gripper embodiments. The OpenXR HAND API outputs a 27-landmark skeletal model \cite{meta2024oxrhand}. Our VR App converts these data into a 21-joint MediaPipe-style topology \cite{lugaresi2020mediapipe}, which our retargeting module consumes and converts to embodiment-specific action (see Fig.~\ref{fig:teleop_pipeline}). At 30\,Hz, the system outputs the wrist pose as a position and quaternion orientation (together representing $T_{\text{wrist},t}\in SE(3)$) and a set of 21 hand landmarks, where each landmark is a 3D position ($\mathcal{K}_t = \{\mathbf{k}_i \in \mathbb{R}^3\}_{i=1}^{21}$).

\begin{figure}[h]
    \centering
    \includegraphics[width=0.45\textwidth]{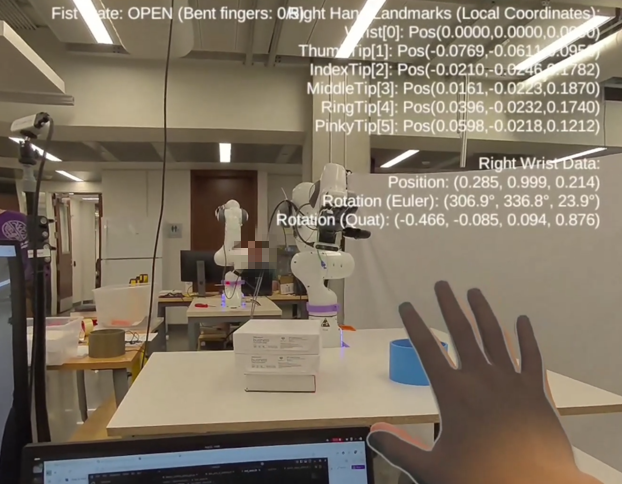}
    \caption{VR Hand-tracking Interface: HUD and shadow overlay showing the tracked operator hand in real time.}
    \label{fig:vr_interface}
\end{figure}

For operator feedback, we render a virtual head-up display (HUD) and an animated overlay that shadows the user's motion in real time. This visualization allows the operator to intuitively verify the motion-capture quality and ensure proper hand alignment during demonstrations, as shown in Figure~\ref{fig:vr_interface}.

To guarantee a reliable data stream for high-quality demonstrations, we use a wired TCP connection over Android Debug Bridge (ADB), which avoids the jitter common to wireless protocols. The 30\,Hz update rate is synchronized with the scene-camera frame rate and policy control loop to minimize motion artifacts and ensure temporal consistency.


\subsection{LeRobot Extension}
\label{sec:lerobot_extension}
We integrate the VR interface with the learning stack through a modular LeRobot extension. Because LeRobot does not natively accept VR inputs, our extension provides an adapter layer that translates wrist-pose and hand-skeleton tracking data streams into the teleoperator abstractions expected by LeRobot, bridging one VR input source to many robot-specific class instantiations. The extension exposes two parallel retargeting branches, one for the positioning of the robot's end-effector, and the other for the positioning of the gripper's fingers. These branches use standard robot kinematic descriptions, such as URDFs, to produce robot actions; simplifying the process of supporting new hardware. The same APIs support both demonstration collection and policy rollout.

\paragraph{Arm Branch - Differential Intent Inverse Kinematics}
\label{sec:arm_branch}
At the start of teleoperation, we record the initial operator wrist pose $T_{\text{wrist},0}$ and robot end-effector pose $T_{\text{ee},0}$.
Each step $t$ computes a differential intent in the operator frame, where $\mathbf{p}(T)\!\in\!\mathbb{R}^3$ denotes the position of pose $T$ in Cartesian space and $\mathbf{q}(T)\!\in\!\mathbb{S}^3$ its unit-quaternion orientation; $\otimes$ is quaternion multiplication:
\begin{align}
\Delta \mathbf{p}_{\text{op}} &= \mathbf{p}(T_{\text{wrist},t}) - \mathbf{p}(T_{\text{wrist},0}), \\
\Delta \mathbf{q}_{\text{op}} &= \mathbf{q}_{\text{wrist},t} \otimes \mathbf{q}_{\text{wrist},0}^{-1}, \quad \text{(quaternion increment).}
\end{align}
A robot-specific calibrated transform $T^{\text{base}}_{\text{op}}$ maps this intent to the robot base, yielding the target end-effector pose:
\[
T^{\text{target}}_{\text{ee}} \;=\; \big(T^{\text{base}}_{\text{op}} [\Delta \mathbf{p}_{\text{op}},\,\Delta \mathbf{q}_{\text{op}}] (T^{\text{base}}_{\text{op}})^{-1}\big)\, T_{\text{ee},0}.
\]
where $[\Delta \mathbf{p},\Delta \mathbf{q}]$ is the affine transform with rotation $R(\Delta \mathbf{q})$ and translation $\Delta \mathbf{p}$.
An IK solver (analytical or numerical) then uses the arm URDF to compute joint targets $\mathbf{q}^{\text{arm}}_t$ while honoring joint limits.

\paragraph{Hand Branch - Dexterous Retargeting}
\label{sec:hand_branch}
The 21-keypoint hand skeleton is normalized and passed to a dexterous retargeting optimizer\cite{qin2023anyteleop} using a DexPilot-style objective\cite{handa2020dexpilot}. This solver is configured using the \emph{hand URDF}, including its joint limits and any mimic joint constraints. The optimization objective is to preserve salient inter-finger geometry from the human demonstration:

\begin{align*}
\min_{\mathbf{q}^{\text{hand}}_t} & \;\sum_i w_i\,\ell\!\big(\mathbf{v}_i^{\text{robot}}(\mathbf{q}^{\text{hand}}_t)-\mathbf{v}_i^{\text{human}}(\mathcal{K}_t)\big) \\
& \text{s.t.}\quad \mathbf{q}_{\min}\!\le\!\mathbf{q}^{\text{hand}}_t\!\le\!\mathbf{q}_{\max},
\end{align*}

where $\mathbf{v}_i^{\text{human}}(\mathcal{K}_t)\!\in\!\mathbb{R}^3$ selects keypoint $i$ from the human skeleton, $\mathbf{v}_i^{\text{robot}}(\mathbf{q})\!\in\!\mathbb{R}^3$ is the corresponding point from forward kinematics at joint state $\mathbf{q}$, $w_i\!\ge\!0$ are saliency weights (e.g., fingertips), and $\ell(\cdot)$ is a robust loss (e.g., Huber). The optimizer outputs joint targets $\mathbf{q}^{\text{hand}}_t$ as robot action commands, with an optional exponential moving average (EMA) filter available to further smooth trajectory execution.

\paragraph{LeRobot Integration}
The LeRobot extension provides a standardized data interface to the learning stack, abstracting away low-level hardware details:
\begin{itemize}
    \item \textbf{Observations} $\mathbf{o}_t$: Synchronized robot proprioception and vision streams.
    \item \textbf{Actions} $\mathbf{a}_t$: A unified structure for joint targets $\{\mathbf{q}^{\text{arm}}_t,\mathbf{q}^{\text{hand}}_t\}$.
\end{itemize}
This consistent schema means that while the extension requires \textbf{embodiment-specific configuration} (e.g., loading the correct URDFs and controller interfaces), the logged datasets and the policy's observation-action interface remain unchanged when swapping robot hardware.


\subsection{Optional Robot I/O Server}
\label{sec:io_server}

In some embodiments, native SDKs already provide direct support for sending commands and accessing proprioceptive feedback. In such cases, the robot can interface with the LeRobot robot classes without additional components. When this interface is not available, we implement a lightweight, embodiment-specific Robot I/O server. Its sole function is to transmit actions and receive proprioceptive data, and it is designed to integrate seamlessly with the \emph{robot class implementation} of the LeRobot framework.


\section{LeFranX System Implementation}

We developed LeFranX as an open-source realization of our modular architecture on a Franka FER robot equipped with a RobotEra XHand. This open-source instantiation serves as a concrete example of the framework's adaptability and consists of the core LeRobot extension, a real-time control server, and a data-to-policy pipeline.

\subsection{The LeFranX Extension}
The LeFranX extension is the embodiment-specific instantiation of the LeRobot extension component of our framework (see Sec~\ref{sec:lerobot_extension}). It consumes the 30\,Hz stream of wrist poses and 21-point hand skeletons from the VR interface and uses the two-branch teleoperation pipeline described in Sec~\ref{sec:lerobot_extension}, with parameters specific to the Franka FER robot and XHand. This same pipeline enables both demonstration collection and policy rollout on the real system.

\paragraph{Robot Arm Teleoperation}
The customization point for the arm-branch of differential-intent inverse kinematics described in Sec~\ref{sec:arm_branch} involves choosing a kinematics solver plugin and scaling factor. For the Franka, the scaling is 1 because it's workspace and movements are close to that of a human arm.  We use the GeoFIK \cite{lopezcustodio2025geofik} solver, a closed-form analytical IK solver specific to the Franka to obtain fast, deterministic solutions ideal for real-time teleoperation. To handle redundancy, we exploit the null space of the 7-DOF arm and optimize the seventh joint with a weighted objective that balances proximity to a neutral configuration, continuity with the current pose, and manipulability. This approach yields accurate, smooth, and stable motion during demonstrations. Full mathematical details are provided in Appendix~\ref{appendix:ik}.

\paragraph{Dexterous Hand Retargeting}
For the RobotEra XHand, we instantiate the hand-branch described in Sec.~\ref{sec:hand_branch} with embodiment-specific parameters. The framework’s hand-level scaling is set to $1\times$, as the XHand’s overall dimensions are close to those of a human hand. To smooth trajectories, we selected an EMA constant of $\alpha = 0.6$. In addition, we introduce a small adjustment for the pinky finger: because it is disproportionately longer than its human counterpart, we apply an adaptive scaling heuristic to preserve natural motion. The resulting 12-DOF joint commands respect joint limits, temporal smoothness, and coupled motions, and are streamed to the hand controller in real time (Fig.~\ref{fig:retargeting_pipeline}). Full mathematical details are provided in Appendix~\ref{appendix:retargeting}.

\begin{figure}[b]
    \centering
    \includegraphics[width=0.475\textwidth]{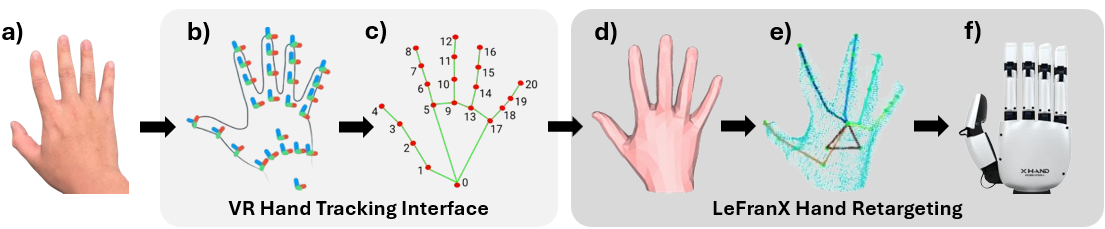}
    \caption{Hand retargeting pipeline: a) operator hand input, b) fit to OpenXR Hand skeleton c) convert to MediaPipe convention, d) convert to MANO convention, e) dex-retargeting with DexPilot optimizer, f) send robot action}
    \label{fig:retargeting_pipeline}
\end{figure}

\paragraph{Assembling the Arm–Hand System}
Since each embodiment is represented as a robot class within the LeRobot framework, the Franka arm and RobotEra XHand can be combined by instantiating a composite robot that encapsulates the individual classes. This modular design makes it straightforward to mix and match different grippers and arms while maintaining a uniform interface for teleoperation, data collection, and policy rollout.

\subsection{Franka Server}
The Franka robot required the implementation of the optional Robot I/O Server (Sec.~\ref{sec:io_server} because the Franka Control Interface (FCI) and `libfranka` C++ library \cite{libfranka} need a real-time Linux kernel, which we run on a separate computer from the main PC used for VR and policy inference.

The server receives high-level action commands from the client and streams back proprioceptive observations, such as joint positions and velocities over TCP. A key responsibility is to bridge the 30\,Hz input command rate with the FCI's strict 1\,kHz control loop. To meet this constraint, we integrate Ruckig \cite{berscheid2021jerk}, an online trajectory generator. Each incoming action serves as a new target for Ruckig, which then generates a smooth, jerk-limited trajectory that is sampled at 1\,kHz to produce continuous, dynamically feasible joint velocity commands for the robot.

\subsection{Demonstration Collection}

We used the LeFranX system to capture expert demonstrations for training and dissemination on Hugging Face\cite{huggingface}, an open-source platform and community for hosting and sharing machine learning models and datasets. To ensure efficient storage, proprioceptive states and actions were recorded in the Apache Parquet format, while high-volume visual data was encoded as compressed MP4 videos with synchronized timestamps \cite{cadene2024lerobot}.

\begin{figure}[b]
    \centering
    \includegraphics[width=0.44\textwidth]{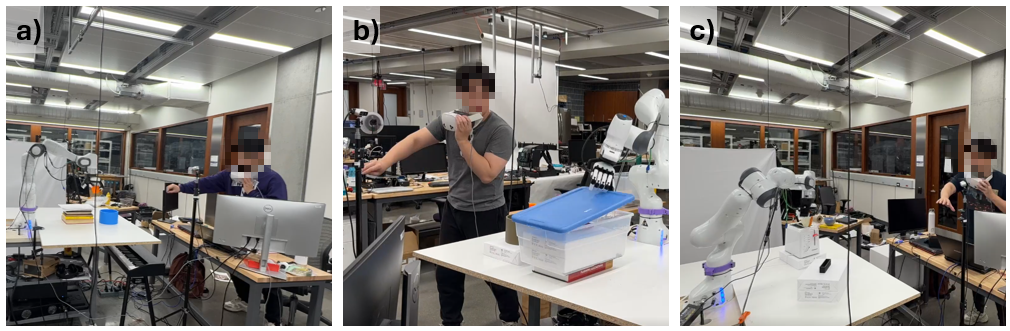}
    \caption{Operator performing demonstrations for the three different tasks: a) pick and place an orange cube into blue bin, b) open the lid, pick the pie in the box and place into brown bin, c) pick up a slice of bread, insert into toaster and push down toaster lever.}
    \label{fig:expert_demonstration}
\end{figure}

The 19-dimensional action space (7-DOF arm, 12-DOF hand) and the observation space, composed of robot proprioception and video from three external Intel RealSense D435 cameras, were synchronized and recorded at 30\,Hz. We collected a dataset across three distinct tasks of increasing complexity (Figure~\ref{fig:expert_demonstration}). For each task, we gathered 100 demonstration episodes.

\subsection{Policy Finetuning and Rollout}

Using the LeRobot framework, we directly finetuned two state-of-the-art visuomotor policies, ACT and DP, on our LeFranX-collected dataset. Each model was finetuned on the 100 expert demonstrations per task for 100k steps, with the best validation checkpoint selected for evaluation. For rollout, the selected models were deployed on the same hardware, receiving synchronized inputs and outputting 19-DOF actions at 30\,Hz. Each task-policy pair was evaluated over 10 rollouts, and success rates were recorded.


\section{Experiment Evaluation}

\begin{figure*}[t]
    \centering
    \includegraphics[width=1.0\textwidth]{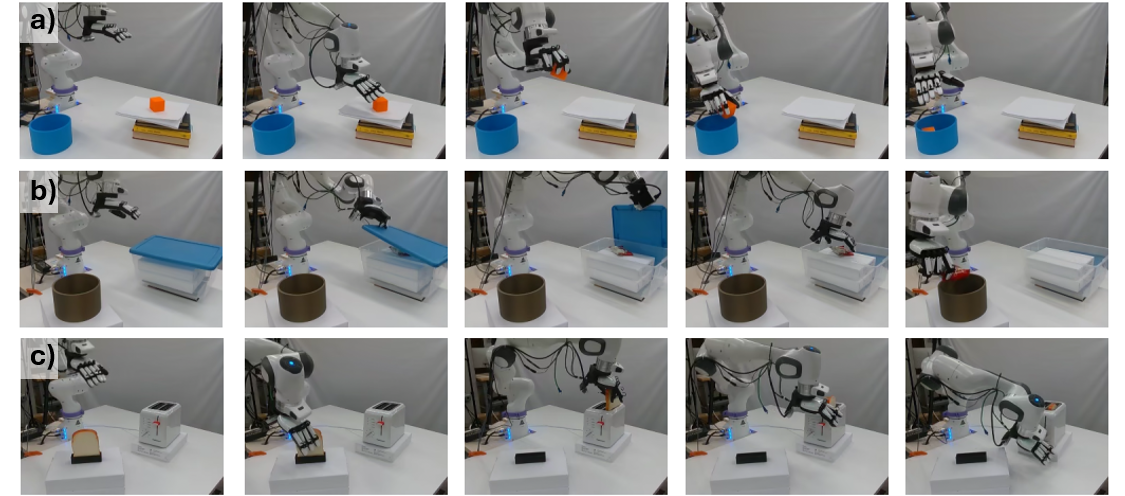}
    \caption{Task sequences for teleoperation and imitation learning. 
    a) Orange Cube Task: pick up the orange cube and place in the blue bin, 
    b) Boxed Pie Task: open a closed box lid, pick up the pie and place in a brown bin, 
    c) Bread Toaster Task: pick up a slice of bread, insert into toaster slot, and press down toaster lever. 
    Video for the tasks is available at \url{https://youtu.be/TzlUEWCjQ1M}.}
    \label{fig:task_sequences}
\end{figure*}

To evaluate the capabilities of our system, we designed three tasks with progressively increasing complexity. These tasks, illustrated in Figure~\ref{fig:task_sequences}, were chosen to provide an assessment of the integrated hand-arm system's dexterity, the effectiveness of the framework, and and the quality of the collected demonstration data. The tasks chosen were:

\begin{itemize}
\item \textbf{Orange Cube Task:} A baseline pick-and-place task where the operator grasps an orange cube and places it in a blue bin, evaluating basic grasping and positioning.
\item \textbf{Boxed Pie Task:} A multi-step task involving lid removal, object retrieval, and placement in a brown bin, testing dexterous manipulation and precise pick-and-place.
\item \textbf{Bread Toaster Task:} The most complex task, requiring grasping a bread slice, inserting it into a narrow slot, and pressing the toaster lever, assessing fine control and sequential actions.
\end{itemize}
\subsection{Setup}

\textbf{Hardware:} Experiments were conducted using a Franka FER robot equipped with a RobotEra XHand dexterous hand. Policy inference was performed on a remote server featuring an NVIDIA RTX 6000 Ada Generation GPU, while a local System 76 workstation managed robot control and data transfer. We used a gRPC-based policy server framework provided by LeRobot for remote policy inference.

\textbf{Vision System:} The visual setup consisted of three Intel RealSense D435 cameras: one overhead, one third-person view, and one wrist-mounted. All cameras provided 320x240 pixel images at 30fps.

\begin{table}[b]
\centering
\small
\caption{Average task completion time (s)}
\label{tab:task_completion_time}
\begin{tabular*}{\columnwidth}{@{\extracolsep{\fill}}lccc}
\toprule
\textbf{System} & \textbf{\shortstack{ Orange\\Cube Task}} & \textbf{\shortstack{Boxed\\Pie Task}} & \textbf{\shortstack{Bread\\Toaster Task}} \\
\midrule
Direct manipulation & 3.2 & 5.7 & 5.5 \\
LeFranX teleoperation  & 6.3 & 11.3 & 12.0 \\
Open-Teach teleoperation & 11.9 & -- & 20.5 \\
\bottomrule
\end{tabular*}
\end{table}

\subsection{Teleoperation Efficiency} 

An effective teleoperation system should be intuitive, minimizing operator cognitive load and enabling efficient task execution. To examine this in our setup, we conducted a case study with a single experienced operator (one of the authors), providing an anecdotal but controlled benchmark for quantifying efficiency. The operator performed each task both directly by hand and via the teleoperation system, with 10 repetitions per condition to obtain stable averages. The results, presented in Table~\ref{tab:task_completion_time}, show that in this setting our system introduced only a modest time overhead, suggesting that complex tasks can be executed efficiently through the VR teleoperation pipeline. 

 \subsection{Policy Evaluation} 
 
 To validate that our pipeline produces high-quality data for imitation learning, we used the collected datasets to finetune two pre-trained visuomotor policies: ACT and DP, as hosted by LeRobot. For each of the three tasks, we finetuned both models using the 100 corresponding expert demonstrations. We then evaluated the performance of each trained policy by executing 10 autonomous rollouts from randomized initial states. A rollout was marked as successful if the agent completed the entire task. The resulting success rates, shown in Table~\ref{tab:policy_success_rates}, demonstrate the efficacy of our collected data for training capable policies. 

\begin{table}[b]
\centering
\small
\caption{Success rate of ACT and DP policies across three tasks}
\label{tab:policy_success_rates}
\begin{tabular*}{\columnwidth}{@{\extracolsep{\fill}}lccc}
\toprule
\textbf{Policy} & \textbf{\shortstack{Orange\\Cube Task}} & \textbf{\shortstack{Boxed\\Pie Task}} & \textbf{\shortstack{Bread\\Toaster Task}} \\
\midrule
ACT & 8/10 & 5/10 & 4/10 \\
DP  & 6/10 & 3/10 & 1/10 \\
\bottomrule
\end{tabular*}
\end{table}

\section{Discussion}

Our results suggest that the proposed VR teleoperation and IL framework is effective for training competent visuomotor policies, yet they also surface several limitations in both teleoperation and policy performance—particularly as task complexity increases. Below, we analyze these challenges and outline promising directions for improvement.

\subsection{Teleoperation Performance}
The results reported here should be viewed as a single-operator case study rather than a definitive comparison between LeFranX and Open-Teach\cite{iyer2024open}. Still, our experience highlights how differences in teleoperation design can affect task performance. In these trials, we found that LeFranX enabled faster task completion than the baseline (Table~\ref{tab:task_completion_time}), largely because its dexterous retargeting preserved the geometric intent of motion, in contrast to the baseline’s direct joint angle mapping. This difference also shaped task feasibility: for example, the baseline could not execute the prying motion required for the Boxed Pie Task. While these observations point to clear advantages of the LeFranX approach, further studies with multiple operators and tasks are needed for a more general assessment.

Even with these improvements, we observed that teleoperation performance still lagged behind direct human manipulation, with teleoperated tasks typically taking about twice as long. In our experience, this gap arose from several factors, including kinematic mismatches between human and robot, inertia-limited differences in maximum speed and acceleration, and the absence of haptic feedback and egocentric vision, which together increased cognitive load. These limitations suggest that future work should explore ways of reducing operator burden, such as integrating richer sensory feedback and adaptive control strategies.

\subsection{Learned Policy Performance}
A key advantage of the LeFranX framework is that it enables direct evaluation of different policies under a uniform teleoperation and deployment pipeline. As shown in Table~\ref{tab:policy_success_rates}, we were able to compare ACT and DP on the same suite of tasks without altering the underlying data collection or control setup. Both policies achieved high success rates on the simpler orange cube pick-and-place task (80\% and 60\%, respectively), while performance degraded on more complex, contact-rich tasks such as bread toaster insertion. 

In running these trials, we also noted differences in how the two policies interacted with the hardware. Diffusion Policy was more sensitive to inference latency than ACT, particularly when executed locally on a workstation with an RTX 4060 GPU. Offloading inference to a remote GPU server reduced but did not eliminate this issue, and the resulting temporal misalignment between observation and action was a recurring source of failure in dynamic tasks.  

By providing a standardized pipeline, LeFranX enables a controlled evaluation of different policies, allowing for a better understanding of the critical interplay between algorithmic design, hardware latency, and the demands of complex tasks.

\subsection{Limitations and Future Work}

A key limitation of our current setup is the tracking latency introduced by the consumer-grade VR hardware. While these devices promote accessibility, their software overhead and network dependence impose delays that affect real-time control. Future work may explore alternative solutions such as optimized network configurations or dedicated FPGA-accelerated tracking and feedback hardware, though these approaches may reduce accessibility due to specialized equipment. More broadly, addressing operator burden—including latency, perceptual mismatches, and limited feedback—remains an important direction for improving both teleoperation fluency and downstream policy performance.


\section{Conclusion}

We introduced LeVR, a modular and generalizable VR teleoperation framework designed for direct integration with modern robot learning pipelines. We also presented \emph{LeFranX}, an open-source implementation of this framework for the Franka Research robot and RobotEra XHand, and released an accompanying public dataset of manipulation demonstrations. Our experiments validated the system's effectiveness, demonstrating low-latency teleoperation and the successful use of our collected data to finetune state-of-the-art visuomotor policies. By contributing a principled framework, a robust implementation, and a ready-to-use dataset, our work provides the research community with a powerful platform to accelerate progress in dexterous manipulation and imitation learning through LeVR.


\section{Acknowledgment}

The authors would like to thank the Northwestern Master of Science in Robotics program for supporting this project, and the Northwestern MAGICS Lab for providing access to the XHand robot.

\bibliographystyle{IEEEtran}
\bibliography{References}


\appendices

\section{Inverse Kinematics Formulation}
\label{appendix:ik}

\subsection{Notation and Initialization}
We denote the VR/world frame by $\{\mathcal V\}$ and the robot base frame by $\{\mathcal B\}$. End-effector poses are $T\in SE(3)$ with position $\mathbf p\in\mathbb R^3$ and orientation represented by quaternion $\mathbf q\in\mathbb H$ or rotation $\mathbf R\in SO(3)$. At teleoperation start we record the operator wrist pose $T_{\text{wrist},0}$ in $\{\mathcal V\}$ and the robot end-effector pose $T_{\text{ee},0}$ in $\{\mathcal B\}$. Because the Franka arm’s workspace is similar to a human arm, we apply unit scaling ($1\times$) to Cartesian increments.

\subsection{Target Pose Computation}
At time step $i$, wrist position increment in the VR frame is
\begin{equation}
\mathbf{p}'_{\mathcal V} = \mathbf{p}_{\mathcal V,i} - \mathbf{p}_{\mathcal V,0}.
\end{equation}
The mapping from VR to robot coordinates is given by
\begin{equation}
\mathbf{T}_{\mathcal V \rightarrow \mathcal B} =
\begin{bmatrix}
0 & 0 & 1 \\
-1 & 0 & 0 \\
0 & 1 & 0
\end{bmatrix}.
\end{equation}
Thus
\begin{equation}
\mathbf{p}'_{\mathcal B} = \mathbf{T}_{\mathcal V \rightarrow \mathcal B}\,\mathbf{p}'_{\mathcal V},\quad
\mathbf{p}_{\text{target}} = \mathbf{p}_{\text{ee},0} + \mathbf{p}'_{\mathcal B}.
\end{equation}
For orientation, we compute the relative quaternion
\begin{equation}
\mathbf{q}_{\delta,\mathcal V} = \mathbf{q}_{\mathcal V,i}\otimes \mathbf{q}_{\mathcal V,0}^{-1},
\end{equation}
map it to the robot frame by
\begin{equation}
\mathbf{R}_{\delta,\mathcal B} = \mathbf{T}_{\mathcal V \rightarrow \mathcal B}\,\mathbf{R}_{\delta,\mathcal V}\,\mathbf{T}_{\mathcal V \rightarrow \mathcal B}^\top,
\end{equation}
and form $\mathbf{q}_{\text{target}} = \mathcal{Q}(\mathbf{R}_{\delta,\mathcal B}) \otimes \mathbf{q}_{\text{ee},0}$. The target pose is then
\begin{equation}
T_{\text{target}} =
\begin{bmatrix}
\mathbf R(\mathbf q_{\text{target}}) & \mathbf p_{\text{target}}\\
\mathbf 0^\top & 1
\end{bmatrix}.
\end{equation}

\subsection{Analytical Inverse Kinematics (GeoFIK)}
The IK problem seeks $\mathbf q=[q_1,\dots,q_7]^\top$ such that $T_{\text{target}}=f(\mathbf q)$. GeoFIK parameterizes redundancy with $q_7$. For fixed $q_7^\ast$, the remaining 6-DOF system yields up to eight solutions:
\begin{equation}
\mathcal S(q_7^\ast)=\{\mathbf q_1,\dots,\mathbf q_k\},\quad k\le8,
\end{equation}
computed via geometric decomposition of shoulder (joints 1–3), elbow (joint 4), and wrist (joints 5–7). Solutions violating joint limits or with low manipulability are discarded, and continuity is enforced by selecting candidates close to $\mathbf q_{t-1}$.

\subsection{Redundancy Resolution}
We optimize over $q_7\in[q_{7,\min},q_{7,\max}]$ using Brent’s method (tol $10^{-6}$, max 100 iters), initialized from $q_{7,t-1}$, with objective
\begin{equation}
J(\mathbf q)=w_m M(\mathbf q)-w_n\|\mathbf W_n(\mathbf q-\mathbf q_{\text{neutral}})\|-w_c\|\mathbf W_c(\mathbf q-\mathbf q_{t-1})\|,
\end{equation}
where $M(\mathbf q)=\sqrt{\det(\mathbf J\mathbf J^\top)}$ is Yoshikawa manipulability. The scalar weights $w_m,w_n,w_c$ trade off manipulability, proximity to a neutral posture $\mathbf q_{\text{neutral}}$, and continuity with the previous state $\mathbf q_{t-1}$. The diagonal matrices $\mathbf W_n,\mathbf W_c$ normalize joint ranges and allow joint-specific emphasis. This scalar search is equivalent to optimizing in the arm’s null space.

\section{Dexterous Hand Retargeting}
\label{appendix:retargeting}

\subsection{Formulation}
Let $\mathbf q_t\in\mathbb R^{12}$ denote the XHand joint angles at time $t$. From VR landmarks we construct $N$ reference vectors $\{\mathbf v_i^{\text{ref}}\}$ (wrist–finger and fingertip–fingertip). For a configuration $\mathbf q_t$, the corresponding robot vectors are
\begin{equation}
\mathbf v_i^{\text{robot}}(\mathbf q_t)=f_i(\mathbf q_t).
\end{equation}
The optimization seeks
\begin{equation}
\min_{\mathbf q_t}\ \sum_{i=1}^N w_i\,\rho_\delta\!\big(\|\mathbf v_i^{\text{robot}}-\mathbf v_i^{\text{ref}}\|\big)+
\lambda\|\mathbf q_t-\mathbf q_{t-1}\|^2
\end{equation}
subject to $\mathbf q_{\min}\le\mathbf q_t\le\mathbf q_{\max}$. Here $\rho_\delta$ is the Huber loss, and $\lambda$ controls temporal smoothing.

\subsection{Adaptive Reference and Weights}
Reference vectors are modified based on landmark proximity:
\begin{equation}
\mathbf v_i^{\text{ref}}=
\begin{cases}
\dfrac{\mathbf v_i^{\text{human}}}{\|\mathbf v_i^{\text{human}}\|}\cdot \eta_j & \|\mathbf v_i^{\text{human}}\|<d_{\text{proj}},\\
\mathbf v_i^{\text{human}}\cdot s & \|\mathbf v_i^{\text{human}}\|>d_{\text{esc}},\\
\text{prev. state} & \text{otherwise},
\end{cases}
\end{equation}
with $d_{\text{proj}}=d_{\text{esc}}=0.03$\,m. Here $\eta_j$ is a projected distance ($\eta_1=10^{-4}$\,m for finger–finger, $\eta_2=3\cdot10^{-2}$\,m for wrist–finger), and $s=1.0$ is a scaling factor applied when landmarks are well separated. Weights are $400$ (finger–finger, projected), $200$ (wrist–finger, projected), and $1$ (free motion).

\subsection{XHand-Specific Adaptations}
Because the XHand dimensions are close to the human hand, we use unit scaling ($1\times$) for overall normalization. To smooth trajectories we apply an exponential moving average
\begin{equation}
\tilde{\mathbf q}_t=\alpha\,\mathbf q_t+(1-\alpha)\,\tilde{\mathbf q}_{t-1},\quad \alpha=0.6.
\end{equation}
Additionally, we further apply an adaptive scaling
$\mathbf v_{\text{pinky}}^{\text{ref}} \leftarrow \gamma_{\text{pinky}}\,\mathbf v_{\text{pinky}}^{\text{ref}}$,
where $\gamma_{\text{pinky}}\in[1.2,2.2]$ increases with the MCP--TIP extension ratio, compensating for the proportionally longer gripper pinky finger.

\subsection{Solver}
Gradients are obtained via the robot Jacobian, and the problem is solved with SLSQP (tol $10^{-6}$, max 200 iters), enforcing joint and coupling limits as hard constraints.

\end{document}